\documentclass[10pt,twocolumn,letterpaper]{article}

\usepackage{cvpr} 

\usepackage{times}
\usepackage{epsfig}

\usepackage{microtype}
\usepackage{graphicx}
\usepackage{subfigure}
\usepackage{booktabs} 

\usepackage{caption}
\usepackage{bbm}
\usepackage{multirow}
\usepackage{multicol}
\usepackage{wrapfig}
\usepackage{nicefrac, xfrac}

\usepackage{arydshln}

\newcommand{\deflen}[2]{%
    \expandafter\newlength\csname #1\endcsname
    \expandafter\setlength\csname #1\endcsname{#2}%
}

\deflen{plotwidth}{200pt}
\makeatletter
\renewcommand{\paragraph}{%
  \@startsection{paragraph}{4}%
  {\z@}{1ex \@plus 1ex \@minus .2ex}{-1em}%
  {\normalfont\normalsize\bfseries}%
}
\makeatother
\usepackage{hyperref}

\title{CHOSEN: Contrastive Hypothesis Selection for Multi-View Depth Refinement}

\author{Di Qiu
\\Google XR \and
Yinda Zhang
\\Google XR \and
Thabo Beeler
\\Google XR \and
Vladimir Tankovich 
\thanks{work done at Google} \and
Christian H\"ane \thanks{work done at Google}
\\Meta Reality Labs \and
Sean Fanello
\\Google XR \and
Christoph Rhemann
\\Google XR \and
Sergio Orts Escolano
\\Google XR \\
}

\begin{document}

\maketitle
\begin{abstract}
  We propose CHOSEN, a simple yet flexible, robust and effective multi-view depth refinement framework.
It can provide significant improvement in depth and normal quality, and can be integrated in existing multi-view stereo pipelines with minimal modifications.
Given an initial depth estimation, CHOSEN iteratively re-samples and selects the best hypotheses.
The key to our approach is the application of contrastive learning in an appropriate solution space and a carefully designed hypothesis feature, based on which positive and negative hypotheses can be effectively distinguished.
We integrated CHOSEN in a basic multi-view stereo pipeline, and show that it can deliver impressive quality in terms of depth and normal accuracy compared to many other top deep learning based multi-view stereo pipelines.

\end{abstract}

\section{Introduction}
Geometry acquisition through multi-view imagery is a crucial task in 3D computer vision.
In the multi-view stereo matching (MVS) framework, image patches or features are matched and triangulated to find 3D points \cite{collins1996space, Furukawa2010}.
The dense geometry is usually represented as a depth map for each view.
Due to view dependent appearance, occlusion and self-similarity in the real world scenarios, the matching signal is often too noisy to directly give accurate and complete geometry.
Traditionally, this problem have been attacked by some carefully designed optimization and filtering scheme \cite{barron2016fast,taniai1603continuous}, directly on the depth map or indirectly inside a matching cost volume, where they impose certain priors on the smoothness of the resulting geometry, making use of the confidence of matching and appearance in a larger spatial context.

\begin{figure}
    \centering
    \includegraphics[width=0.5\textwidth]{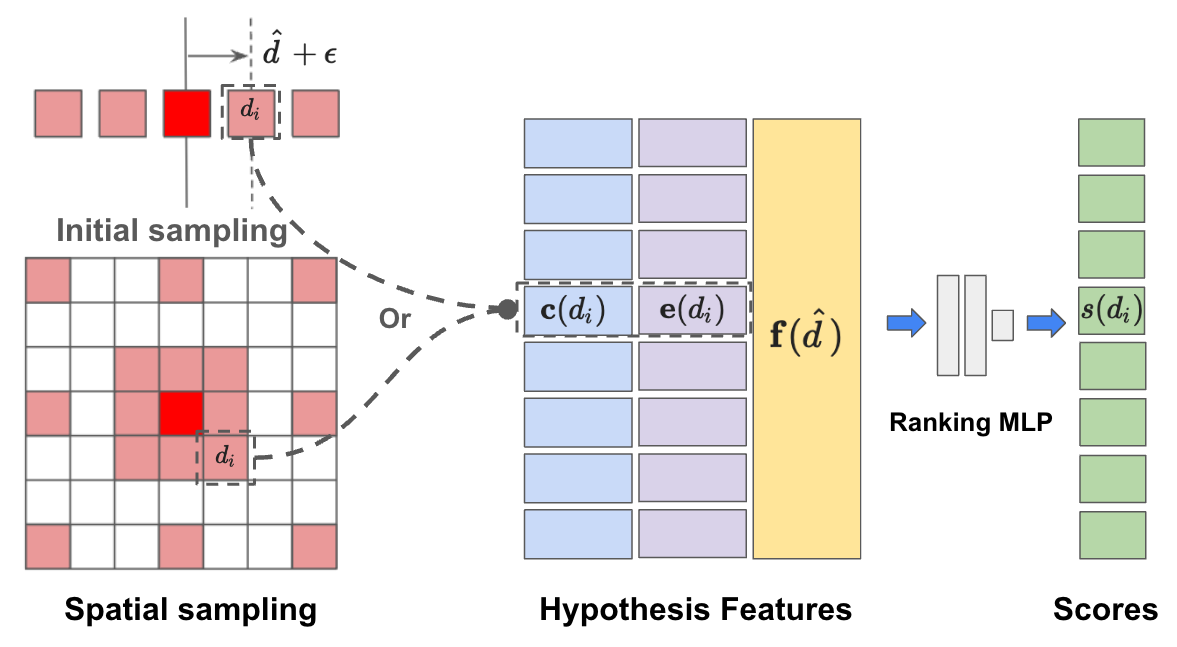}
    \caption{CHOSEN's hypotheses sampling and ranking mechanism. Assuming an initial depth estimation, we gather depth hypotheses from its perturbations or its spatial neighbors. For each hypothesis, we combine the matching cost, a second order smoothness term and a context feature to form the input of a learned score function represented by an MLP. The one with the highest score will be selected to update refined depth. The process is performed iteratively.}
    \label{fig:core}
\end{figure}

With the advancement of deep learning, many research works have combined the ideas with traditional MVS.
Early works \cite{yao2018mvsnet,YaoYao2019RMVSNet} typically involve extracting convolutional features from the images, building one cost volume (indexed by a sequence monotone depth hypotheses $d_i$ for each pixel), or several in a multi-resolution hierarchy, and transform these cost volumes using 3D convolutions into a "probability volume" so that depth map can be obtained by taking the expectation
\begin{equation} \label{eq:expectation}
\hat{d} = \sum_{i} p_i d_i
\end{equation}
More recent works have more complicated designs regarding cost computation and aggregation \cite{wang2021patchmatchnet}, visibility and uncertainty estimation \cite{zhang2023vis,cheng2020ucsnet}, loss functions \cite{peng2022rethinking}, feature backbones \cite{caomvsformer}, feature fusion \cite{zhe2023geomvsnet}, and consequently have yielded impressive progress on various benchmarks.
The evaluation has been limited to point cloud reconstruction, and little attention was paid to, and thus the lack thereof, the raw output's quality in terms of depth and normal accuracy.
Looking closely at the predicted depth and its associated normals, we can often find quantization-like artifacts.
Moreover, the fixed training and testing benchmarks also result in limited discussion on the learning and inference task under the variance of the multi-view acquisition setups, including different metric scales, camera relative positioning and lenses (focal lengths). Therefore, for a new capture setup substantially different from the training set, one may need to determine what hyper-parameters to use, especially the number of depth proposals to sample from a certain depth range, and retrain the model of choice to get the optimal performance.

To some extent, the problem can be attributed to the approach of using Eq.\eqref{eq:expectation} to obtain depth estimations: the distribution of depths hypotheses at different pixels, or across different datasets, can be arbitrary.
It is thus difficult to guarantee the learned probability weights to produce consistent results: in a smooth spatial neighborhood, small perturbations in the ${d_i}$ results in non-smoothness in the predicted depth and normal; across different datasets, the inappropriately spaced ${d_i}$ that are too sparse or too coarse results in poor generalization.

In this work, we turn our attention to the problem of learning to refine depth with multi-view images while keeping in mind that they can be from very different setups, and how to best sample hypotheses for depth refinement.
We will discuss in Sec.\ref{sec:representation} the proper solution space to perform the learning task, and present an alternative to the approach of Eq.\eqref{eq:expectation} in Sec.\ref{sec:ranking}.
Much as inspired by the PatchMatch \cite{Barnes:2009:PAR,bleyer2011patchmatch} methodology, our framework, named CHOSEN, learns to refine the depth by iteratively re-sampling and selecting the best hypotheses, thus eliminating the need of a "probability volume".

In the following we identify and underline the core design elements that enable CHOSEN to produce significantly improved depth and normal quality:
\begin{itemize}
    \item Transformation of the depth hypotheses into a solution space defined by the acquisition setup.
    \item The use of first order approximation to sample spatial hypotheses to propagate good hypotheses to their neighborhood.
    \item A carefully designed hypothesis feature so that it's informative for contrastive learning.
    \item Each hypothesis is evaluated independently so that the refinement is robust to arbitrary hypothesis samples.
\end{itemize}
An overview of CHOSEN is illustrated in Fig.\ref{fig:core}.

To demonstrate the effectiveness of CHOSEN, in Sec.\ref{sec:baseline} we describe how we embed it in a minimalist MVS pipeline where the only extra learnable part is the feature extraction using light-weight U-Nets \cite{unet}. 
We perform comprehensive ablation experiments to justify our design, and compare the quality of our refined depths with various recent deep learning based MVS pipelines.
Without bells and whistles, our baseline model is easy to train, fast to converge, and already delivers impressive depth estimation quality.

\section{Related work}

MVS algorithms are often categorized by the representation used to reconstruct the output scene, e.g. volume \cite{KutulakosS2000, Kar2017, Ji2017SurfaceNet}, point cloud \cite{Furukawa2010, Lhuillier2005, Lin2018, Insafutdinov2018}, depth map \cite{galliani2016gipuma, Tola2012, Schonberger2016}, etc. However, depth map is probably the most flexible and efficient representation among existing ones. While depth map can be considered as a particular case of point cloud representation, (e.g., pixel-wise point), it reduces the reconstruction task to a per-view (2D) depth map estimation. These MVS approaches can be further grouped in hand-crafted (traditional) methods or learning based solutions. Traditional MVS pipelines extend the two-view case \cite{scharstein2002taxonomy} by introducing a view-selection step that aggregates the cost from multiple images to a given reference view. The view selection can be performed per-camera \cite{Galliani15} or per-pixel \cite{Schonberger2016}. These approaches \cite{Galliani15,galliani2016gipuma,Tola2012,Schonberger2016} then rely on well engineered photometric cost functions (ZNCC, SSD, SAD, etc.) to estimate the scene 3D geometry, by selecting the best depth hypothesis that leads to the lowest aggregated cost. However these cost functions usually perform poorly on texture-less or occluded areas, and under complex lighting environments where photometric consistency is unreliable. Hence further post-processing and propagation steps are used to improve the final estimate \cite{Schonberger2016,Galliani15}. We refer readers to \cite{Furukawa2015Tutorial} for additional details regarding traditional multi-view stereo pipelines. 

Recently, MVS algorithms have showed impressive quality of 3D reconstructions in terms of accuracy and completeness, mostly thanks to the increase popularity of deep learning based solutions \cite{yao2018mvsnet, YaoYao2019RMVSNet, Yu2019FastMVSNet, Gu2020CasMVSNet, wang2021patchmatchnet, zhe2023geomvsnet, caomvsformer}. These methods  often make use of multi-scale feature extractors \cite{cheng2020ucsnet, tankovich2021hitnet,chang2018pyramid}, cost-volumes \cite{kendall2017end,stereonet,activestereonet}, and guided refinement \cite{yao2018mvsnet,pang2017cascade,stereonet} to retrieve the final 3D estimate. Typically, they leverage U-Nets \cite{unet} to build a single or a hierarchy of cost volumes with predetermined depth hypotheses. Then, these cost volumes are regularized using 3D convolutions and the final depth map is regressed from the regularized probability volume. However, to achieve high resolution depth accuracy, it requires sampling a large number of depth planes, which is limited by memory consumption.

Researchers are also opening other frontiers in deep learning based MVS.
UCS-Net \cite{cheng2020ucsnet} and Vis-MVSNet \cite{zhang2023vis} use some uncertainty estimate for an adaptive generation of cost volumes. PVSNet \cite{Xu2020PVSNet} and PatchmatchNet \cite{wang2021patchmatchnet} learn to predict visibility for each source image.
The approach of Eq.\eqref{eq:expectation} is considered critically in \cite{peng2022rethinking}, although it was only considered from the loss function perspective.
GeoMVSNet \cite{zhe2023geomvsnet} proposes a geometry fusion mechanism in the MVS pipline.
Most recently, transformer based methods \cite{ding2022transmvsnet,caomvsformer,cao2024mvsformer++} exploits the attention mechanism for more robust matching and context awareness.
In particular, MVSFormer \cite{caomvsformer} has combined this approach with the powerful pre-trained DINO features \cite{oquab2023dinov2}.

It is worth mentioning that existing methods usually need to hand-tune hyper-parameters such as the depth range, hypotheses spacing, and number of hypotheses to ensure sufficient and accurate coverage for the new application.
Some of these methods adopt 2D convolutional neural networks to obtain final depth estimations, using RGB image to guide depth up-sampling and refinement \cite{stereonet, Hui2016MSG-Net, wang2021patchmatchnet}. Consequently, these methods often generalize poorly to new camera setups or new scenes.


\section{Framework}

There are two important aspects in our CHOSEN framework.
In Sec.\ref{sec:representation} we discuss how to define a suitable solution space for the depth hypotheses and how to sample them.
This has to accommodate the fact that input depths can be of arbitrary scale, and are computed from different multi-view systems.
Then in Sec.\ref{sec:ranking} we detail our design of a ranking module that is able to process arbitrarily sampled hypotheses, and how it can be trained to effectively and robustly distinguish the good hypotheses from the bad ones.


\subsection{Hypothesis representation and sampling} \label{sec:representation}

\paragraph{Pseudo disparity representation}
We find it crucial to operate in a transformed inverse depth representation, which we call pseudo disparity - akin to disparity for rectified stereo pairs.
Denoting the metric depth as $D$, the pseudo disparity writes 
\begin{equation}
d = \frac{f * b}{D}
\end{equation}
where we choose $f$ to be the focal length (unit in pixels) of the reference camera, and $b$ the metric distance between the reference camera and the {\it closest} source camera.
The scaling factor $f * b$ converts the inverse depth into the {\it pixel space}, and approximates the correct accuracy level of the capture setup in use, granted that the cameras have sufficient frustum overlap and similar focal lengths.
This representation not only allows us to build hypothesis feature insensitive to the variance in metric or intrinsic scales, but also defines the correct granularity in the solution space where we distinguish the positive and negative samples for contrastive learning, thus enabling our depth refinement model to learn and infer across various acquisition setups.

\paragraph{Initial hypotheses sampling}
The initial depth hypotheses are constructed as a $H\times W \times N$ volume where we will look for the optimal hypothesis per pixel that has the lowest matching cost.
In case there is no initial depth available, we initialize the volume using uniformly spaced slices in the range $[d_{\text{min}}, d_{\text{max}}]$, where the number of slices can be chosen so that the slices sit $1\text{-px}$ apart from each other, and initialize the depth $\hat{d}$ to be the one with lowest matching cost.
We refer to the corresponding cost volume as the {\bf full cost volume}.
In case an previous depth $\hat{d}$ is available, the volume will be restricted to be uniformly spaced in $[\hat{d} - M, \hat{d} + M]$.
A cost volume built using this set of hypotheses is referred to as a {\bf local cost volume}, and $\hat{d}$ will be updated to be the one with lowest matching cost in the local volume.
In addition, we apply uniform random perturbation in $\mathcal{U}[-0.5, 0.5]$ to each hypothesis at each pixel for both volumes to robustify the training and estimation.
Note that there are no other particular restrictions on choices of $N$ or $M$ as long as the spacing is appropriate and the coverage is sufficient, and in particular they can be changed without re-training.


\begin{figure*}[t!]
    \centering
    \includegraphics[width=1.0\textwidth]{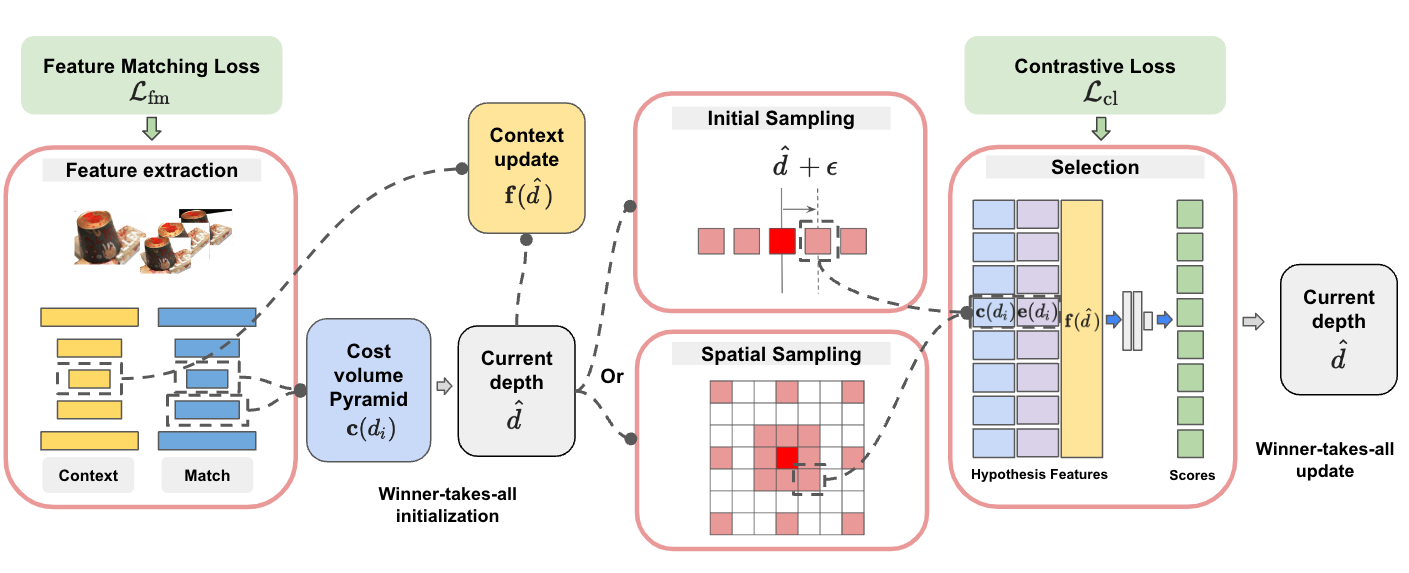}
    \caption{Overview of the baseline MVS with our CHOSEN depth refinement. The winner hypothesis is from the initial full range cost volume, followed by applying the hypotheses sampling and best hypothesis selection. Initial sampling and spatial sampling are applied in an alternating fashion. Spatial sampling is facilitated through first order propagation as defined in Eq.\eqref{eq:first order}. Key to this pipeline is the design of the hypothesis feature, defined in Sec.\ref{para: hypothesis feature}. The refined depth is upsampled to the higher resolution using nearest neighbor, and the same refinement procedure will be applied.}
    \label{fig:pipeline}
\end{figure*}

\paragraph{Spatial hypothesis sampling}
Inspired by the PatchMatch framework \cite{Barnes:2009:PAR,bleyer2011patchmatch}, we use spatial hypotheses to expand good solutions into their vicinity.
Specifically, we sample a set of hypotheses for each pixel by collecting the {\it propagated} depths from its spatial neighbors. 
The propagation is facilitated through first order approximation. Denote $\partial d = (\partial_x d, \partial_y d)$ to be gradient of the depth $d$, $\mathbf{p}' = \mathbf{p}+\Delta \mathbf{p}$ be a spatial neighbor of the pixel $\mathbf{p}$. The propagated depth from $\mathbf{p}'$ to $\mathbf{p}$ is defined as
\begin{equation} \label{eq:first order}
    d_{\mathbf{p}'\to \mathbf{p}} = d_{\mathbf{p}'} - \partial d_{\mathbf{p}'} \cdot  \Delta \mathbf{p}
\end{equation}
In practice we use a fixed set of $\Delta \mathbf{p}$'s for each pixel, though it is more of a convenience than requirement.
We conduct sampling and best hypothesis selection in multiple iterations and resolutions.

\subsection{Contrastive learning for hypothesis ranking} \label{sec:ranking}
It is crucial to keep in mind that we {\it don't assume any structure in the sampled set of hypotheses}.
In fact, many among them can be wildly bad samples, especially if the initial depth is too noisy.
Hence it implies that each hypothesis should be evaluated independently, and our task is to learn to distinguish the good hypotheses from the bad ones.

To achieve our objective, we designate the ranking model to be a small MLP that takes a {\bf hypothesis feature} and its {\bf context feature}, and outputs a score for the input hypothesis.
Among all the hypotheses at a particular pixel, the one with the highest score will be selected to be the updated depth estimation $\hat{d}$.
The ranking and selection goes on iteratively to refine the depth estimation, where the new $\hat{d}$ will be used to generate better hypotheses with the updated context feature.

Our design for the hypothesis feature and context update mechanism is illustrated in Fig.\ref{fig:core}.
There are three general guidelines we have followed when we design the hypothesis feature: (1) It should inform how well the matching is given by the hypothesis; (2) It should inform how well the hypothesis fits into the current spatial context; (3) It should be robust to different camera setups and resolutions.
With these in mind, we propose to use the concatenation of the following three simple quantities computed from a hypothesis $d_i$:
\begin{equation} \label{eq:feature}
[\mathbf{c}(d_i), \mathbf{e}(d_i), \mathbf{f}(\hat{d})]
\end{equation}
The first term $\mathbf{c}(d_i)$ consists of the matching costs linearly interpolated from the cost volume for a fixed set of perturbations
\begin{equation} \label{eq:feature:c}
    \mathbf{c}(d_i) = [c(d_i + \epsilon) \text{ for } \epsilon \in \{-1, 0, 1\}]
\end{equation}
Note that $\epsilon$ is defined in the pseudo disparity space.

The second term $\mathbf{e}(d_i)$ is a "tamed" version of the second-order error in the one-ring neighborhood $\mathcal{N}$ of the pixel $\mathbf{p}$.
Note that the error is computed against ${\hat{d}}$ in the pseudo disparity space so that it can work across different metric and intrinsic scales.
\begin{equation}
    \mathbf{e}(d_i) = [\text{tanh}(\hat{d}_{\mathbf{p} + \Delta p} - \partial \hat{d}_{\mathbf{p}'} \cdot \Delta p - d_i) \text{ for } \Delta \mathbf{p} \in \mathcal{N}]
\end{equation}

Finally, $\mathbf{f}(\hat{d})$ is a learned context feature of the previous refined depth, whose inputs are the concatenation of $\mathbf{c}(\hat{d})$, $\mathbf{e}(\hat{d})$ and the feature from a U-Net describing the reference view's high level appearance.
We use a two-layer convolution network to initialize $\mathbf{f}(\hat{d})$, and subsequently we use a convolutional gated recurrent unit (GRU) to update $\mathbf{f}(\hat{d})$ given the updated $\hat{d}$.
We remark that the overall iterative update is similar to the RAFT \cite{teed2020raft} framework, with the addition of the geometric error term $\mathbf{e}(d)$. We will discuss its crucial contribution in our ablation study in Sec.\ref{para: hypothesis feature}.

We use a contrastive loss to supervise the ranking module.
Let $d_{\text{gt}}$ be the ground truth depth and ${\mathcal{D}} = {d_i}$ be the set of input hypotheses. We define the positive sample group to be those within $1$-pseudo disparity of the ground truth
\begin{equation}
\mathcal{D}^{+} = \{d_i \in \mathcal{D}: |d_i - d_{\text{gt}}| <= 1\}
\end{equation}
Denote the score of $d_i$ to be $s(d_i)$.
Our contrastive loss is defined as
\begin{equation}
\mathcal{L}_{\text{cl}} = -\log (\sum_{d_i \in \mathcal{D}^{+}}\frac{\exp (s(d_i))}{\sum_{d_i \in \mathcal{D}} \exp(s(d_i))})
\end{equation}
As result, the ranking module will learn to put most of the weight on the set $\mathcal{D}^{+}$.
We found this formulation most effective in training, and much superior compared to learning to put all weight on the closest sample to the ground truth. We believe it's due to that during spatial propagation, many samples will be already close to the ground truth and supervising on the closest sample introduces unnecessary competition among good samples. 

\begin{figure}[t!]
    \centering
    \includegraphics[width=0.5\textwidth]{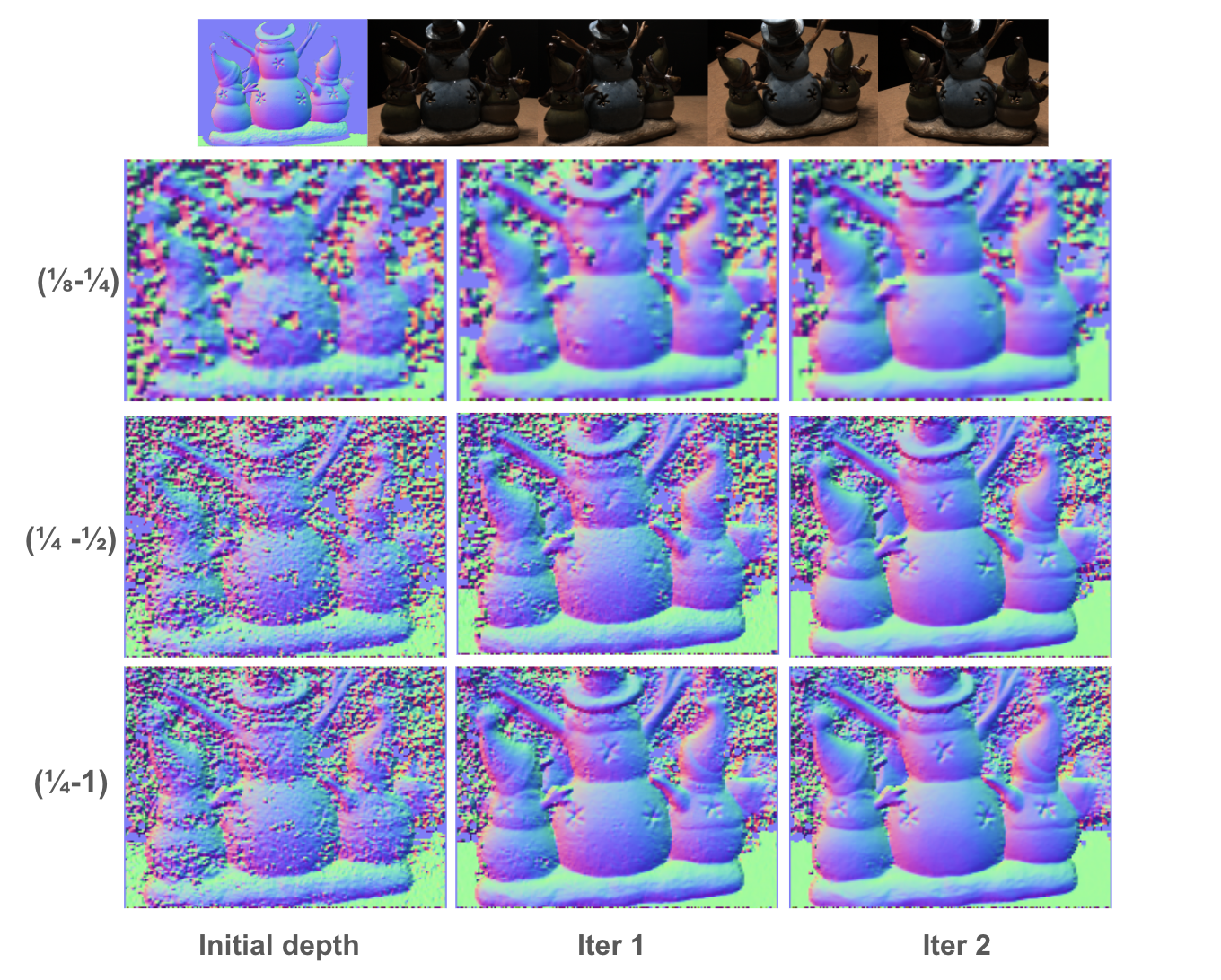}
    \caption{Evolution of the refined depth in our baseline MVS model. The first row shows the ground truth and input images, and we mark each row with the cost volume pyramid configuration. As one can see, the winner-take-all initialization from the cost volume is usually very noisy, but nevertheless contain some accurate estimations. By iteratively re-sample and selecting the best hypotheses, the depth and quality are significantly improved.}
    \label{fig:evolution}
\end{figure}

\subsection{Baseline MVS design} \label{sec:baseline}
Here we will demonstrate the design of a simple end-to-end MVS pipeline with our depth refinement method.
We have deliberately designed it such that {\it there are no learnable components for cost volume construction, depth refinement, or confidence prediction. The only learnable parts are the ranking MLP, context update convolutional layers, the light-weight U-Nets used for extracting matching feature as well as the high level appearance of the reference view}.
Hence our cost volume construction should be inferior and more noisy compared to many existing MVS models that are equipped with cost volume filtering.
Even with our simple pipeline, we show in Sec.\ref{sec:depth comparison} that it can achieve much more superior quality in depth and normal compared to many state-of-art MVS works on the DTU dataset.

\paragraph{Feature extraction}
We use a two-stream feature extraction architecture similar to \cite{teed2020raft}.
First, we use a matching feature network extracts distinctive features that are then used to evaluate the cost of depth hypotheses between the reference and other source views.
Second, we use a context feature network that learns high-level spatial cues for learnable hypothesis selection, applied only on the reference image.
Both networks are based on a light-weight U-Net architecture \cite{unet} which extracts features from coarse to fine.

\paragraph{Matching cost aggregation}
We choose the negative correlation as the matching cost:
\begin{equation}
    c_v(d) = -\text{corr}(\mathbf{f}_{v\to\text{ref}}(d),  \mathbf{f}_{\text{ref}})
\end{equation}
where $\mathbf{f}_{v\to\text{ref}}(d)$ denotes the $v$-th source view feature warped to the reference view through the hypothesis $d$.
We aggregate the costs from all source views to have a single cost volume simply as
\begin{equation} \label{eq: agg_correlation}
c(d) = \frac{1}{\Sigma_v w_v(d)}\Sigma_i w_v(d) c_v(d)
\end{equation}
where
\begin{equation}
    w_{i}(d) = \text{Sigmoid}( \alpha(\delta-c_i(d) )^3)
\end{equation} computes the weighting for each view and hypothesis, and $\alpha$ is chosen so that it downplays the contribution from higher cost, increases the importance of lower cost, and not sensitive if the cost is close to $\delta$, thanks to the cubic exponent.

\paragraph{Cost volume pyramid}
As an optional benefit, we observe that it's possible to compute the matching costs for lower resolution depth using higher resolution feature.
Specifically, we construct a grid ${(x_p, y_p)}_p$, indexed by lower resolution pixels, which samples the corresponding locations in the higher resolution feature. We can then compute the warped pixel location and the matching costs using the camera intrinsics and features corresponding to the higher resolution. 

In practice, we typically compute one additional cost volume using higher resolution features.
The new hypotheses are centered around the hypothesis of lowest cost in the coarse cost volume, spaced according the scale defined by the higher resolution pseudo disparity space.
These two cost volumes now form a {\bf cost volume pyramid}, which will be used to provide both coarse and fine matching information to the our ranking module through $\mathbf{c}(d)$ in Eq.\eqref{eq:feature:c}.
We will demonstrate in Sec.\ref{para:cost volume} that using the higher resolution feature can unblock a higher level of accuracy even at lower resolution output, but it's otherwise {\it not} crucial to the success of our hypothesis ranking module.

\paragraph{Overall pipeline}
Our baseline MVS pipeline, as depicted in Figure \ref{fig:pipeline}, starts by extracting features from a reference view and multiple source views.
The first depth estimation $\hat{d}$ is obtained as the lowest cost hypothesis from the full cost volume.
A cost volume pyramid is then constructed around the previous depth estimation.
The refinement is performed on a hierarchy of resolutions.
The refined depth from lower resolution will be upsampled using nearest-neighbor and set to be the initial depth for the next higher resolution.

\paragraph{Training}

Similar to the contrastive loss used in \cite{tankovich2021hitnet}, we supervise the feature matching using a simple contrastive loss
\begin{equation}
    \mathcal{L}_{\text{fm}} = c(d_{\text{gt}}) - \text{clip}(c(d^{-}), [\alpha, \beta])
\end{equation}
where $c(d_{\text{gt}})$ is the ground truth depth, and $d^{-}$ is the negative sample that is at least $1$-pseudo disparity away from the ground truth and has the lowest matching cost
\begin{equation}
    d^{-} = \arg\min_{|d - d_{\text{gt}}| > 1} c(d)
\end{equation}
The loss will not penalize cost of $d^{-}$ already lower than $\alpha$ or higher than $\beta$.
We choose $\beta = 1$ in lower resolution and $\alpha = 0$ for all resolution, so that the learned features will be as distinctive as possible.
We found it better to choose a lower $\beta$ in higher resolution (e.g. 0.8) due to the frequent appearance of local texture-less regions.
Note that there is no any kind of cost volume regularization.
Furthermore, we enforce the matching feature network to be only trained by the above loss, meaning it will not be influenced by the depth refinement.
Therefore, the total loss of this baseline model is simply
\begin{equation}
    \mathcal{L}_{\text{fm}} + \mathcal{L}_{\text{cl}}
\end{equation}
where ${\mathcal{L}_{\text{fm}}}$ only updates the matching feature U-Net, and ${\mathcal{L}_{\text{cl}}}$ updates the context feature U-Net in the reference view, the ranking MLP and the convolutional layers that initialize and update ${\mathbf{f}(\hat{d})}$ in Eq.\eqref{eq:feature}.

\subsection{Integration in MVS pipeline}
It's clear to see that CHOSEN can be embedded end-to-end in existing deep learning based MVS framework with little modification needed, since CHOSEN only needs the cost, second order error term and context feature to rank the depth hypotheses.
One may substitute our simplistic yet noisy cost volume construction in the baseline with the techniques proposed in recent works, such as cost volume filtering \cite{yao2018mvsnet} and other feature backbones \cite{caomvsformer}, etc.
The main thing to consider for the integration is that the depth hypotheses must be converted to the pseudo disparity representation in order to be processed by CHOSEN. Since the spacing of the hypotheses in the cost volume is crucial for contrastive learning, it might be necessary to retrain the model with CHOSEN from scratch since the original model might be sensitive to different spacing.
In Sec\ref{sec: combine_with_3dcnn} We illustrate the some results demonstrating the further improvement CHOSEN by introducing cost volume filtering based on MVSNet \cite{yao2018mvsnet}.
\section{Experiments}

\begin{table*}[t!]
    \centering
    \scalebox{0.9}{
    \begin{tabular}{|l|c|c|c|c|c|}
        \hline
        Method & $\%<1$mm $\uparrow$ & MAE(@$<1$mm) $\downarrow$ & $\%<5^{\circ}$(@$<1$dsp) $\uparrow$ & $\%<10^{\circ}$(@$<1$dsp) $\uparrow$ \\
        \hline
        baseline: (\nicefrac{1}{8},\nicefrac{1}{4},\nicefrac{1}{4})-(\nicefrac{1}{4},\nicefrac{1}{2},1) & $\mathbf{70.16}$ & $\mathbf{0.3669}$ & $\mathbf{45.53}$  & $\mathbf{73.34}$\\
        baseline: (\nicefrac{1}{8},\nicefrac{1}{4},\nicefrac{1}{4})-(\nicefrac{1}{4},\nicefrac{1}{2},1) + BlendedMVS & $67.01$ & $0.3992$ & $43.77$ & $71.73$ \\
        \hdashline
        baseline: (\nicefrac{1}{8},\nicefrac{1}{8},\nicefrac{1}{8})-(\nicefrac{1}{4},\nicefrac{1}{2},1) & $\mathbf{71.03}$ & $\mathbf{0.3558}$ & $\mathbf{68.16}$ & $\mathbf{84.01}$ \\
        baseline: (\nicefrac{1}{8},\nicefrac{1}{8},\nicefrac{1}{8})-n.a. & $63.62$ & $0.3978$ & $57.31$ & $79.64$ \\
        \hdashline
        baseline w/o selection: (\nicefrac{1}{8},\nicefrac{1}{8},\nicefrac{1}{8})-(\nicefrac{1}{4},\nicefrac{1}{2},1) & $59.33$ & $0.4282$ & $45.38$ & $74.41$\\
        \hdashline
        baseline w/o $\mathbf{e}(d)$ term: (\nicefrac{1}{8},\nicefrac{1}{8},\nicefrac{1}{8})-(\nicefrac{1}{4},\nicefrac{1}{2},1) & $55.26$ & $0.4223$ & $41.51$ & $66.63$ \\
        \hdashline
        baseline w/o CHOSEN: (\nicefrac{1}{8},\nicefrac{1}{8},\nicefrac{1}{8})-(\nicefrac{1}{4},\nicefrac{1}{2},1) & $46.46$ & $0.4653$ & $20.90$ & $49.00$\\
        \hline
        MVSFormer \cite{caomvsformer} & $61.92$ & $0.4248$ & $18.96$ & $44.17$ \\
        GeoMVSNet \cite{zhe2023geomvsnet} & $63.93$ & $0.4089$ & $15.18$ & $38.40$ \\
        GBi-Net \cite{mi2022generalized} & $34.01$ & $0.4531$ & $13.78$ & $33.41$ \\
        IterMVS \cite{wang2021itermvs} & $51.09$ & $0.4524$ & $16.67$ & $37.50$ \\
        PatchMatchNet \cite{wang2021patchmatchnet} & $56.19$ & $0.4260$ & $18.03$ & $41.82$ \\
        UCSNet \cite{Gu2020CasMVSNet} & $59.03$ & $0.3942$ & $31.31$ & $54.39$ \\
        \hline
    \end{tabular}}
    \vspace{5pt}
    \caption{Comparison for estimated depths and normals on DTU testing set. The metric is computed on all the valid ground truth pixels. We mark our best results separately for $\nicefrac{1}{8}$ and $\nicefrac{1}{4}$ resolutions. The results from other methods are evaluated at $\nicefrac{1}{4}$ resolutions with nearest neighbor down-sampling. Notice that so long as the finest resolution matching features are used, the final depth accuracy metrics for these two output resolutions are very similar. Our simple baselines offer significant improvement in terms of depth quality even compared to the strongest state-of-art MVS pipelines.}

\label{tab:depth comparison}
\end{table*}

\subsection{Implementation details}
We evaluate our depth refinement method using the baseline MVS pipeline described in Sec.\ref{sec:baseline}.
Typically, we extract matching features at $\nicefrac{1}{8}$, $\nicefrac{1}{4}$, $\nicefrac{1}{2}$ and $1$ of the original resolution, and context features at $\nicefrac{1}{8}$ and  $\nicefrac{1}{4}$ of the original resolution.
The depth is initialized by winner-take-all from the full cost volume at $\nicefrac{1}{8}$ resolution that has $N=128$ slices, and a cost volume pyramid is built on $\nicefrac{1}{8}\,\&\,\nicefrac{1}{4}$ resolution matching features.
One refinement stage consists of 4 iterations, where twice on initial sampled hypotheses and twice on spatially sampled hypotheses, in an alternating fashion.
After upsampling the refined depth using nearest neighbor, we repeat the same refinement stage at $\nicefrac{1}{4}$ resolution, with a $\nicefrac{1}{4}\,\&\,\nicefrac{1}{2}$ pyramid.
Finally, the last refinement stage is performed with a  $\nicefrac{1}{4}\,\&\,1$ pyramid with output at $\nicefrac{1}{4}$ resolution.
Each refinement stage has their own parameters for the ranking MLP and context update networks.
We denote this default configuration as $(\nicefrac{1}{8},\nicefrac{1}{4},\nicefrac{1}{4})\,\&\,(\nicefrac{1}{4},\nicefrac{1}{2},1)$.
Different choices for cost volume pyramid configurations and their effects will be discussed in Sec\,\ref{para:cost volume}.

We set $M=4$ in the initial hypotheses sampling, giving $2M + 1=9$ hypotheses.
For spatial sampling, we use a set of offsets ${\{\Delta \mathbf{p}\}}$ consists of dilated $3\times3$ regular patches with dilation rate of $1$ and $3$, without repeating the patch center, giving in total $17$ hypotheses for each pixel. This offset configuration is the same as the one illustrated in the spatial sampling part of Fig.\ref{fig:core}.

Our baseline model is trained and tested on a single NVIDIA A100 GPU (40G). The total trainable parameter count is about 1.1 million, including the matching and context U-Nets. Our selection module's learnable parameter count is about 781k. We first train with batch size 4, at input resolution $600\times 800$, up to 200k iterations using the default Adam optimizer \cite{kingma2014adam} at a learning rate of $0.001$. We then fine-tune for up to 50k iterations at input resolution $1200\times1600$ at a learning rate of $0.0005$.
During training, we fix the closest source view and randomly sample two from the remaining source views, which is the only data augmentation we used.
For testing, it takes around $0.6$ seconds to run at $1200\times1600$ with $5$ source views, with a total of 12 iterations of hypotheses ranking (4 iterations for each cost volume config).
We provide results both from models trained on DTU \cite{jensen2014large} training set only, and on DTU mixed with the BlendedMVS \cite{yao2020blendedmvs} dataset, which will be tagged specifically.

\paragraph{Depth \& normal metrics}
We use the following metrics to measure the quality of depth and its induced normal map on the DTU evaluation dataset:
\begin{itemize}
    \item $\%<x$mm: the percentage of pixels that have less than $x$-mm of absolute error.
    \item MAE(@$<x$mm): the mean absolute error on pixels that have less than $x$-mm of absolute error.
    \item $\%<x^{\circ}$(@$<y$-dsp): the percentage of pixels where the normal is within $x^\circ$ of angular error out of all pixels whose absolute error is less than $y$-pseudo disparity.
\end{itemize}
The ground truth depth are resized to the same resolution of the outputs using nearest neighbor.
The normals are computed from the 3D coordinate's gradients using the Sobel filter.
All the above metrics are computed only on the valid pixels where the ground-truth is available.

\begin{figure*}[t!]
    \centering
    \subfigure{\scalebox{0.7}{\raisebox{30pt}{\rotatebox[origin=t]{90}{Color image}}}
    \includegraphics[width=0.24\textwidth]{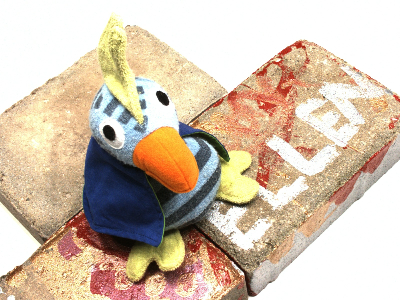}}
    \subfigure{\includegraphics[width=0.24\textwidth]{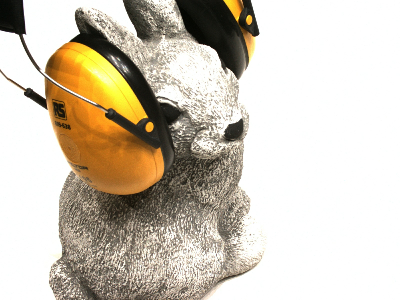}}
    \subfigure{\includegraphics[width=0.23\textwidth]{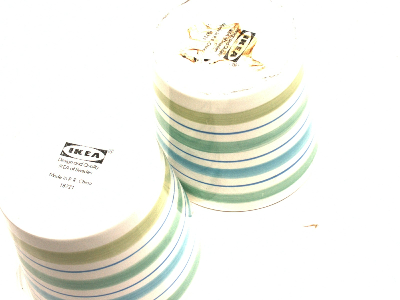}}
    \subfigure{\includegraphics[width=0.23\textwidth]{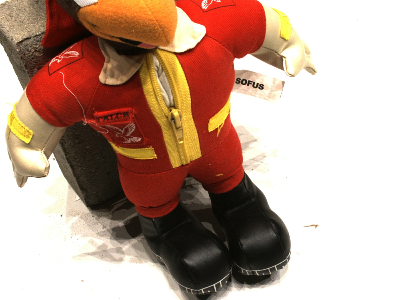}}
    \vspace{-10pt} \\
    \subfigure{\scalebox{0.7}{\raisebox{30pt}{\rotatebox[origin=t]{90}{Ground truth}}}
    \includegraphics[width=0.24\textwidth]{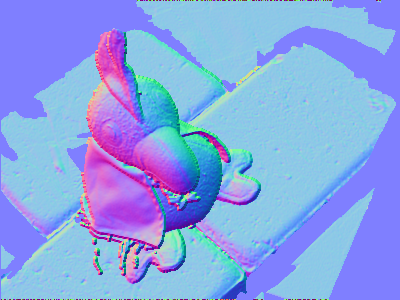}}
    \subfigure{\includegraphics[width=0.24\textwidth]{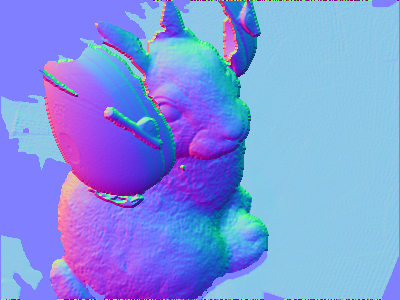}}
    \subfigure{\includegraphics[width=0.24\textwidth]{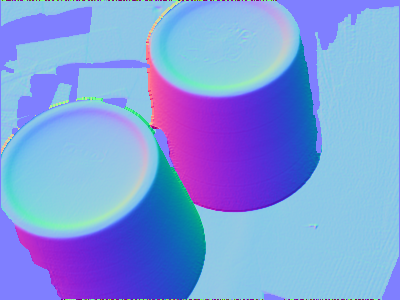}}
    \subfigure{\includegraphics[width=0.24\textwidth]{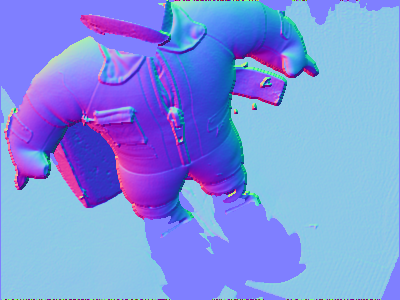}}
    \vspace{-10pt} \\
    \subfigure{\scalebox{0.7}{\raisebox{30pt}{\rotatebox[origin=t]{90}{Ours }}}
    \includegraphics[width=0.24\textwidth]{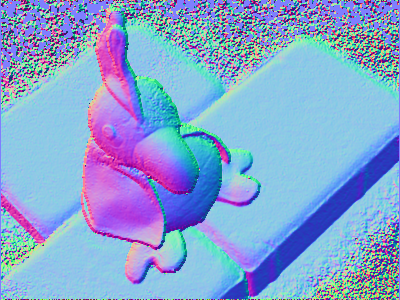}}
    \subfigure{\includegraphics[width=0.24\textwidth]{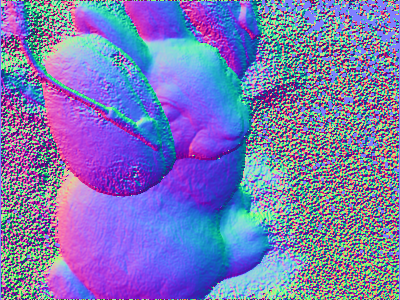}}
    \subfigure{\includegraphics[width=0.24\textwidth]{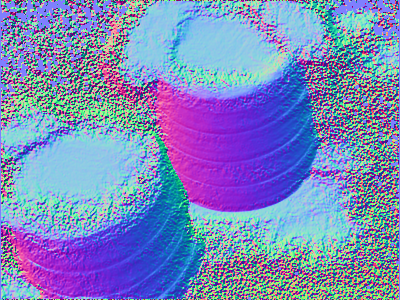}}
    \subfigure{\includegraphics[width=0.24\textwidth]{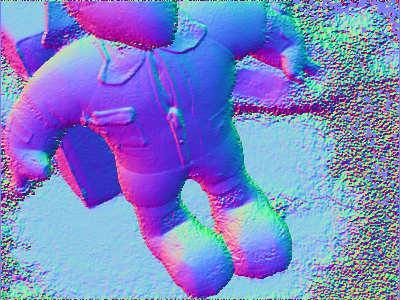}}
    \vspace{-10pt} \\
    \subfigure{\scalebox{0.7}{\raisebox{20pt}{\rotatebox[origin=t]{90}{MVSFormer}}}
    \includegraphics[width=0.24\textwidth]{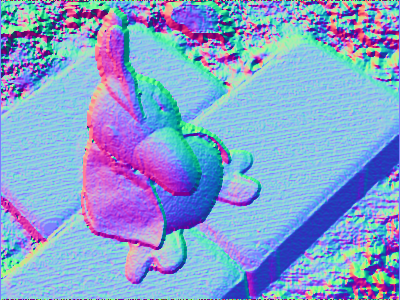}}
    \subfigure{\includegraphics[width=0.24\textwidth]{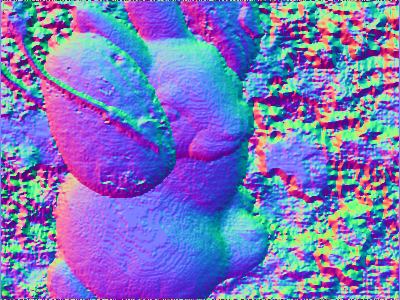}}
    \subfigure{\includegraphics[width=0.24\textwidth]{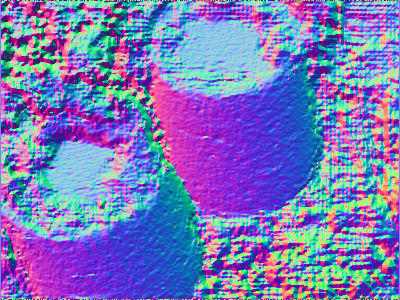}}
    \subfigure{\includegraphics[width=0.24\textwidth]{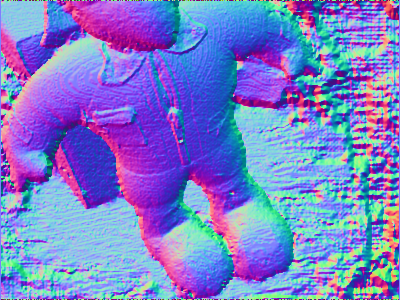}}
    \vspace{-10pt} \\
    \subfigure{\scalebox{0.7}{\raisebox{20pt}{\rotatebox[origin=t]{90}{GeoMVSNet}}}
    \includegraphics[width=0.24\textwidth]{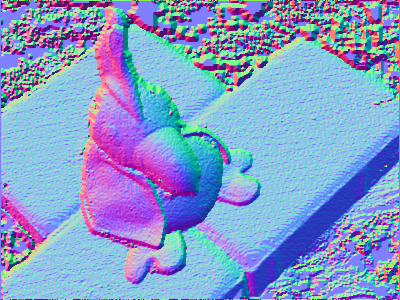}}
    \subfigure{\includegraphics[width=0.24\textwidth]{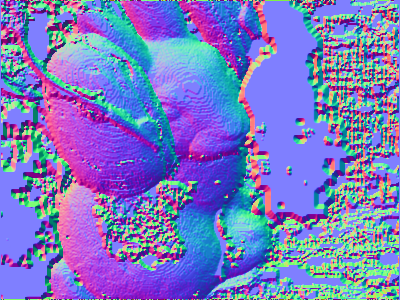}}
    \subfigure{\includegraphics[width=0.24\textwidth]{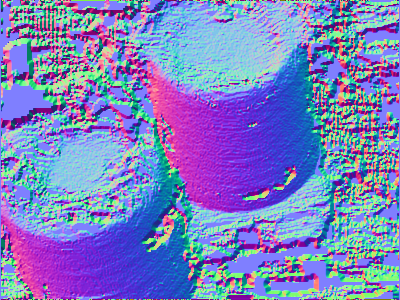}}
    \subfigure{\includegraphics[width=0.24\textwidth]{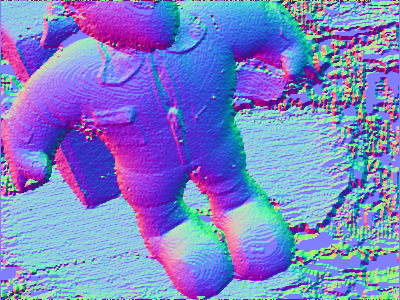}}
    \vspace{-10pt} \\
    \subfigure{\scalebox{0.7}{\raisebox{20pt}{\rotatebox[origin=t]{90}{GBi-Net}}}
    \includegraphics[width=0.24\textwidth]{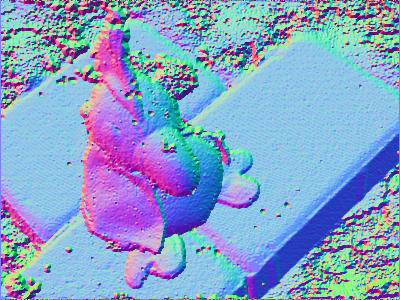}}
    \subfigure{\includegraphics[width=0.24\textwidth]{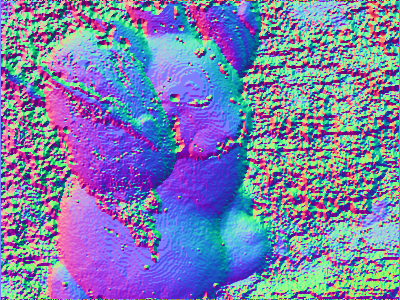}}
    \subfigure{\includegraphics[width=0.24\textwidth]{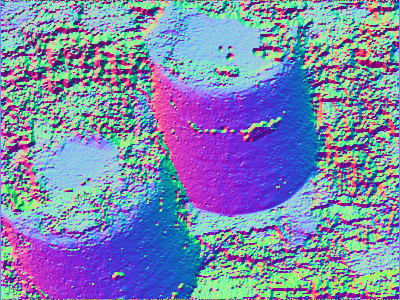}}
    \subfigure{\includegraphics[width=0.24\textwidth]{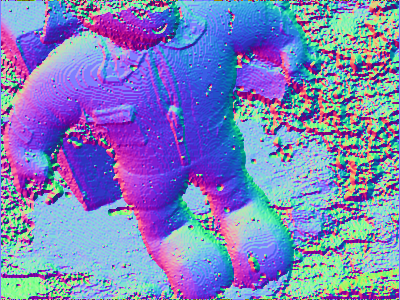}} \vspace{0pt}\\
\caption{Normal quality comparisons on DTU. Our simple baseline MVS trained only on DTU produces significantly more accurate normals.}
\label{fig:dtu_compare}
\end{figure*}

\begin{figure*}[t!]
\centering
\subfigure{\includegraphics[width=0.16\textwidth]{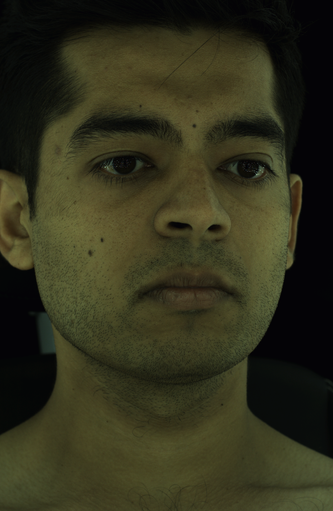}}
\subfigure{\includegraphics[width=0.16\textwidth]{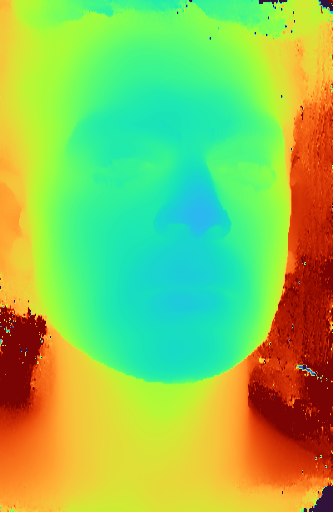}}
\subfigure{\includegraphics[width=0.16\textwidth]{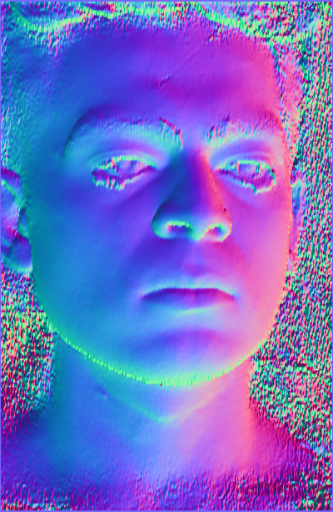}}
\subfigure{\includegraphics[width=0.16\textwidth]{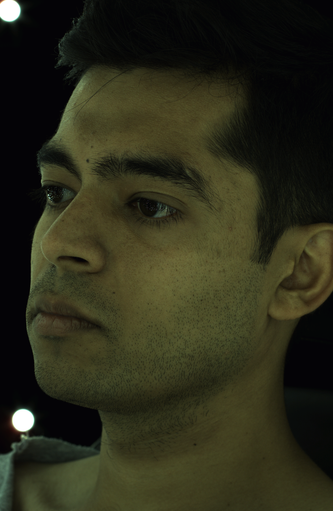}}
\subfigure{\includegraphics[width=0.16\textwidth]{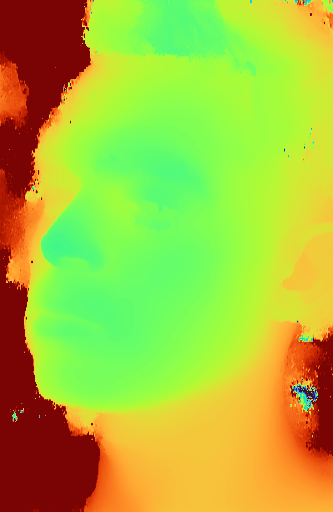}}
\subfigure{\includegraphics[width=0.16\textwidth]{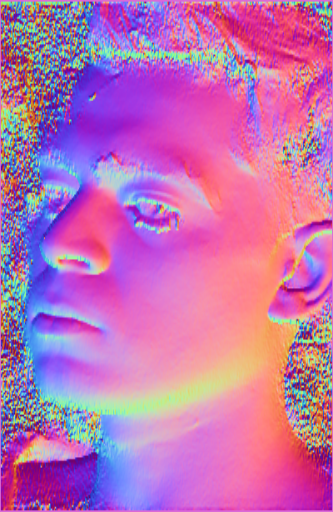}} \\ \vspace{-10pt}
\subfigure{\includegraphics[width=0.16\textwidth]{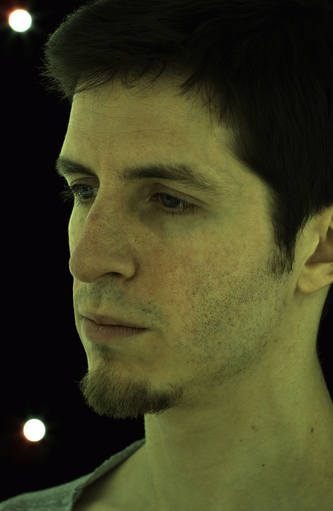}}
\subfigure{\includegraphics[width=0.16\textwidth]{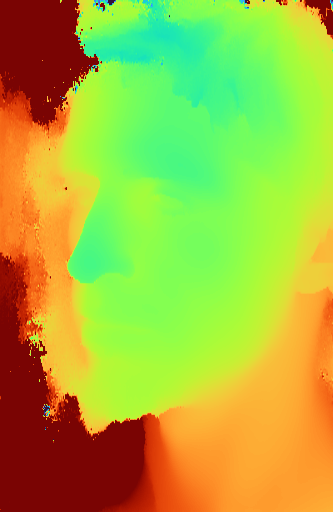}}
\subfigure{\includegraphics[width=0.16\textwidth]{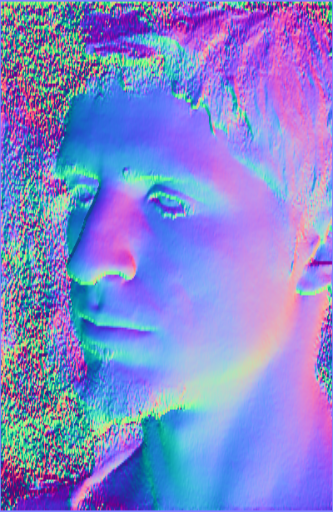}}
\subfigure{\includegraphics[width=0.16\textwidth]{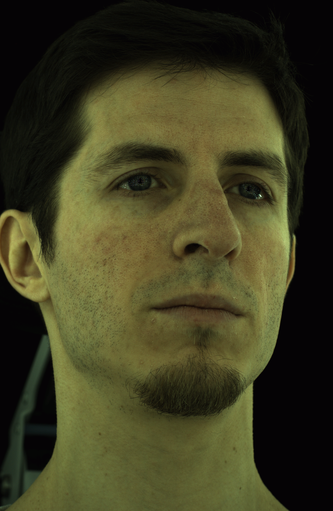}}
\subfigure{\includegraphics[width=0.16\textwidth]{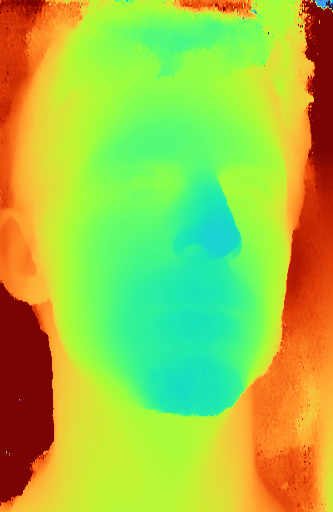}}
\subfigure{\includegraphics[width=0.16\textwidth]{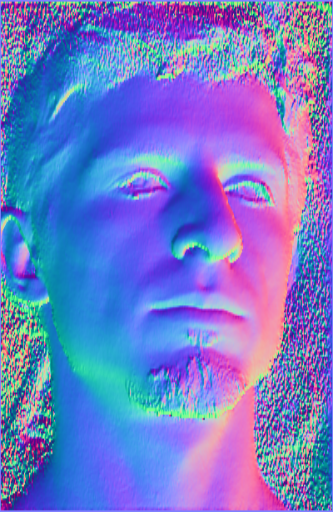}} \\ \vspace{-10pt}
\subfigure{\includegraphics[width=0.16\textwidth]{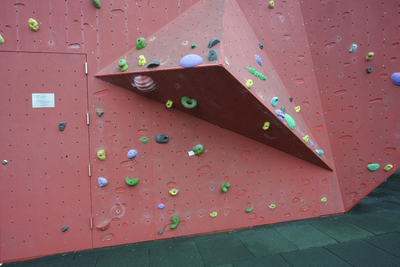}}
\subfigure{\includegraphics[width=0.16\textwidth]{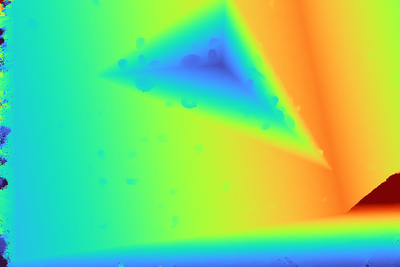}}
\subfigure{\includegraphics[width=0.16\textwidth]{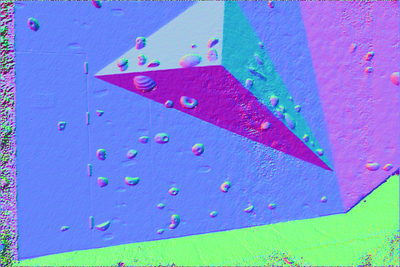}}
\subfigure{\includegraphics[width=0.16\textwidth]{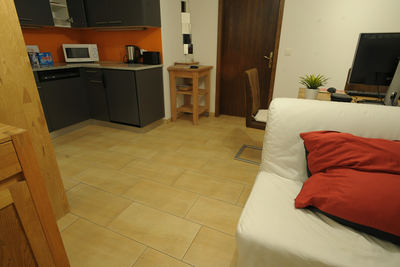}}
\subfigure{\includegraphics[width=0.16\textwidth]{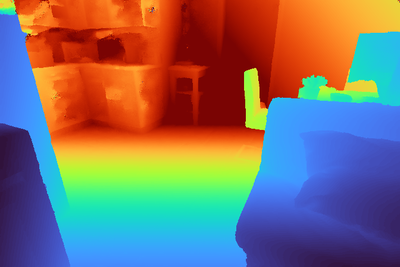}}
\subfigure{\includegraphics[width=0.16\textwidth]{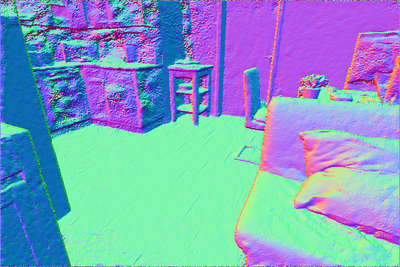}} \\ \vspace{-10pt}
\subfigure{\includegraphics[width=0.16\textwidth]{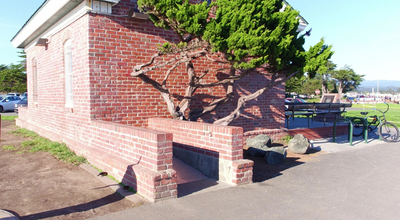}}
\subfigure{\includegraphics[width=0.16\textwidth]{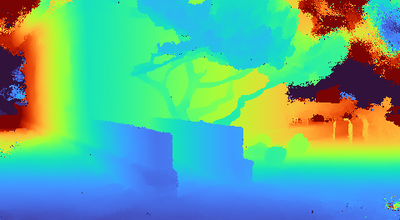}}
\subfigure{\includegraphics[width=0.16\textwidth]{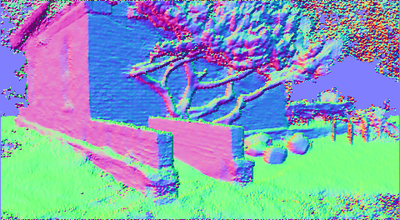}}
\subfigure{\includegraphics[width=0.16\textwidth]{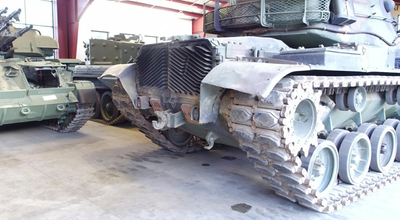}}
\subfigure{\includegraphics[width=0.16\textwidth]{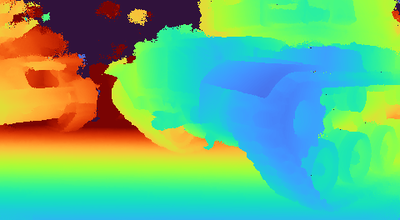}}
\subfigure{\includegraphics[width=0.16\textwidth]{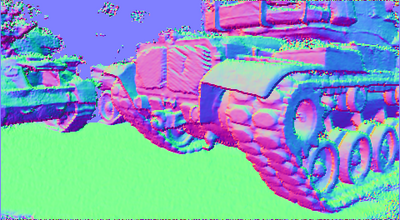}}\\
\caption{Direct application of our DTU + BlendedMVS trained baseline model on instances from MultiFace \cite{knapitsch2017tanks}, Tanks \& Temples and ETH3D \cite{schops2017multi} datasets. Our simple baseline achieves consistent generalization ability even though trained on substantially different data.}
\label{fig:wild test}
\end{figure*}

\begin{figure*}[t!]
\subfigure{\includegraphics[width=0.16\textwidth]{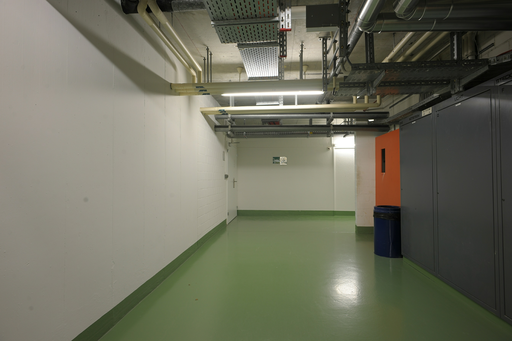}}
\subfigure{\includegraphics[width=0.16\textwidth]{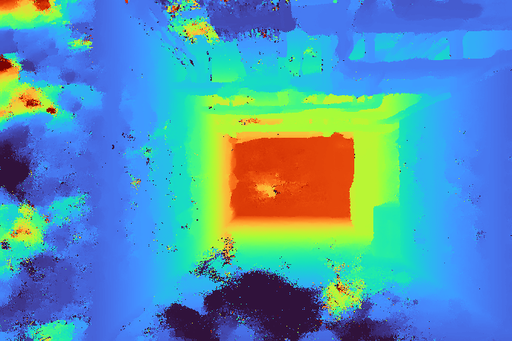}}
\subfigure{\includegraphics[width=0.16\textwidth]{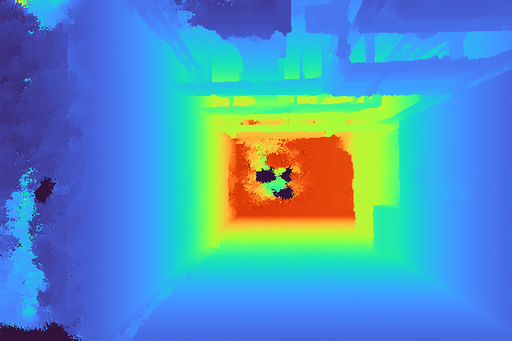}}
\subfigure{\includegraphics[width=0.16\textwidth]{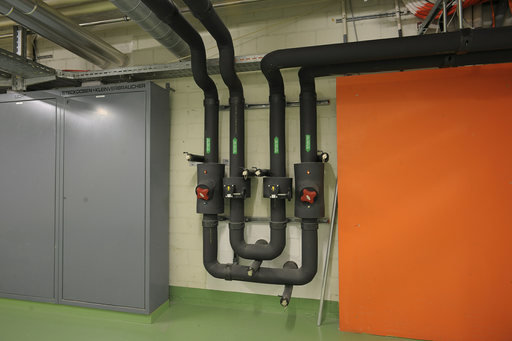}}
\subfigure{\includegraphics[width=0.16\textwidth]{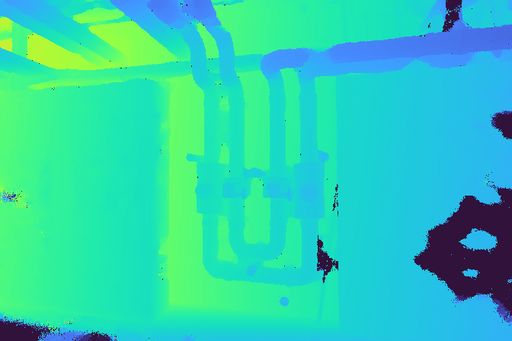}}
\subfigure{\includegraphics[width=0.16\textwidth]{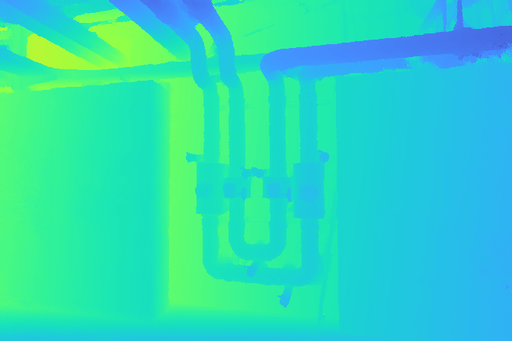}} \vspace{-10pt} \\
\subfigure{\includegraphics[width=0.16\textwidth]{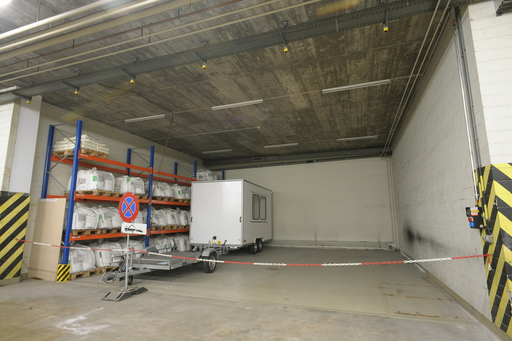}}
\subfigure{\includegraphics[width=0.16\textwidth]{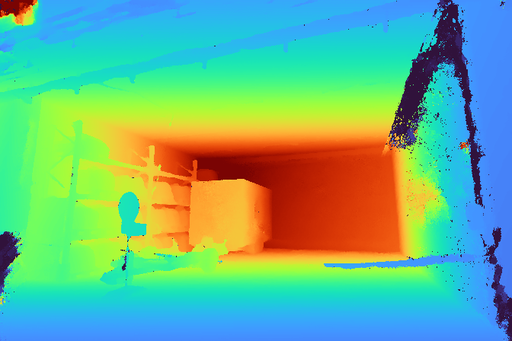}}
\subfigure{\includegraphics[width=0.16\textwidth]{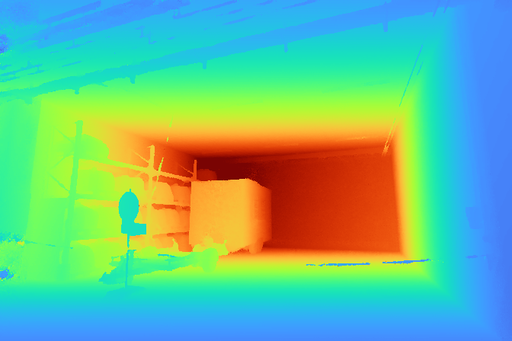}}
\subfigure{\includegraphics[width=0.16\textwidth]{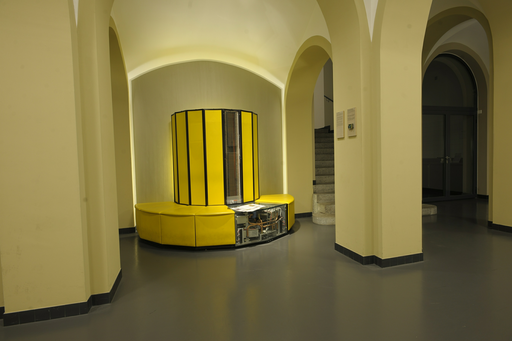}}
\subfigure{\includegraphics[width=0.16\textwidth]{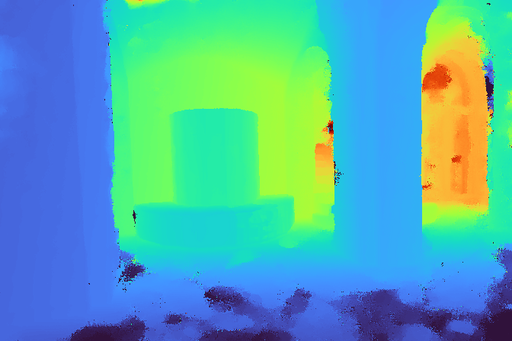}}
\subfigure{\includegraphics[width=0.16\textwidth]{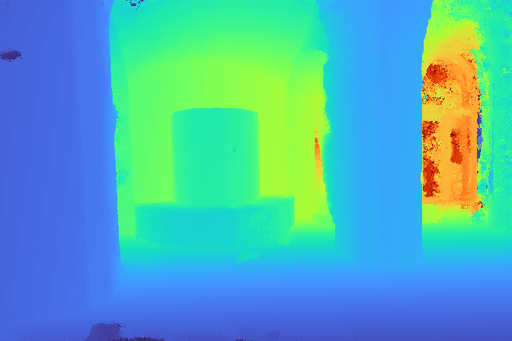}}\\
\caption{From left to right: The reference view, CHOSEN in baseline MVS, and CHOSEN in baseline w/ 3D-CNN. 3D-CNN is applied to the full range cost volume. The improvement is most significant in large texture-less or repetitive regions.}
\label{fig:with_3dcnn}
\end{figure*}

\subsection{Ablation studies}

\paragraph{Cost volume pyramid design} \label{para:cost volume}
Here we study the effects of different cost volume pyramid configurations.
We choose the following variants of the configuration:
\begin{itemize}
    \item ($\nicefrac{1}{8}$,$\nicefrac{1}{4}$,$\nicefrac{1}{4}$)-($\nicefrac{1}{4}$,$\nicefrac{1}{2}$,$1$): There are 3 stages of refinement with $\nicefrac{1}{8}\,\&\,\nicefrac{1}{4}$, $\nicefrac{1}{4}\,\&\,\nicefrac{1}{2}$ and $\nicefrac{1}{4}\,\&\,1$ cost volume pyramids, and output refined depth at $\nicefrac{1}{4}$ resolution. This is our default configuration and we also include the results trained on data mixed BlendedMVS.
    \item ($\nicefrac{1}{8}$,$\nicefrac{1}{8}$,$\nicefrac{1}{8}$)-($\nicefrac{1}{4}$,$\nicefrac{1}{2}$,$1$): There are 3 stages of refinement with $\nicefrac{1}{8}\,\&\,\nicefrac{1}{4}$, $\nicefrac{1}{8}\,\&\,\nicefrac{1}{2}$ and $\nicefrac{1}{8}\,\&\,1$ cost volume pyramids, and output refined depth at $\nicefrac{1}{8}$ resolution.
    \item ($\nicefrac{1}{8}$,$\nicefrac{1}{8}$,$\nicefrac{1}{8}$)-n.a: Only the coarsest resolution matching feature is used. There are 3 stages of refinement with $\nicefrac{1}{8}$ resolution feature, and output refined depth at $\nicefrac{1}{8}$ resolution.
\end{itemize}
Quantitative results on DTU testing set are reported in the first four rows in Tab.\ref{tab:depth comparison}.
First of all, notice that even without cost volume pyramid, our method performs reasonably well, indicating that the pyramid design {\it mainly unblocks higher accuracy at lower resolution output, but otherwise is an optional feature}.
Second, we can observe from the first and third rows in  Tab.\ref{tab:depth comparison} that the $\%<1$mm metrics are similar for the outputs at $\nicefrac{1}{8}$ resolution and $\nicefrac{1}{4}$ resolution, so long as the full resolution matching features are used.
Comparing $\nicefrac{1}{8}$ (third and fourth rows) and 
$\nicefrac{1}{4}$ (first and second rows) output resolution, we can observe that it becomes more difficult to get accurate normals at higher resolution.
This is due to the fact that high frequency details only emerge in higher resolution, which may be impossible to obtain without a photometric appearance model.
Lastly, we remark that the worse results from mixed data training is likely due to that BlendedMVS contains images with severe aliasing artifacts, which could hurt our model training with $\mathcal{L}_{\text{fm}}$.

\paragraph{Selection v.s. expectation}
The ability to rank arbitrary hypotheses is essential to the success of our depth refinement.
Here we illustrate the point by comparing with an approach where the learned score $s_i$ for each hypothesis $d_i$ is used for taking a weighted average. 
In this ablated approach, the refined depth is obtained by
$$
\hat{d} = \frac{1}{\sum_i \exp(s_i)} \sum_i \exp({s_i}) \cdot d_i
$$
and we use smoothed $L^1$ loss between $\hat{d}$ and $d_{\text gt}$ in place of $\mathcal{L}_{\text cl}$.
Everything else in the pipeline stays the same.
Quantitative results are reported in the fifth row of Tab.\ref{tab:depth comparison}.
We note that this alternative approach based on weighted average is much less accurate than the our approach based on ranking and selection.
This can be attributed to the difficult task of learning the weights for arbitrary hypotheses, while learning to {\it classify} the hypotheses is a much easier task.

\paragraph{Hypothesis feature design} \label{para: hypothesis feature}
The "tamed" second order error term $\mathbf{e}(d)$ in Eq.\eqref{eq:feature} can be viewed as an extra component compared to the input feature design in the RAFT \cite{teed2020raft} framework for optical flow estimation.
Here we show that it is a vital component in the hypothesis ranking model that improves the overall smoothness and accuracy.
Quantitative results on DTU testing set are reported in the sixth row of Tab.\ref{tab:depth comparison}.
Since each hypothesis is evaluated independently, without the information about how well a particular hypothesis fits in the local geometry, the model struggles to select the best hypothesis solely based on matching and context information.

\subsection{Comparisons on depths and normals} \label{sec:depth comparison}
We collect the testing results from various recent deep learning based MVS works in the last part of Tab.\ref{tab:depth comparison}.
The outputs for the candidate models and ground truths are resized to $\nicefrac{1}{4}$ of input resolution using nearest neighbor.
Our simple baseline MVS pipeline significantly outperforms the strongest state-of-art in terms depth and normal quality.
Visual comparisons are shown in Fig.\ref{fig:dtu_compare}.

\subsection{Qualitative results for baseline}
We demonstrate the excellent generalization ability of our simple baseline on various datasets including MultiFace \cite{wuu2022multiface}, Tanks \& Temples \cite{knapitsch2017tanks} and ETH3D \cite{schops2017multi}.
Results in Fig.\ref{fig:wild test} use the same baseline model trained on mixture of DTU and BlendedMVS reported in Tab.\ref{tab:depth comparison}.

\subsection{A simple extension of the baseline MVS with cost volume filtering} \label{sec: combine_with_3dcnn}

We explore a very simple extended setting to our baseline MVS, where we apply an MVSNet \cite{yao2018mvsnet} style 3D-CNN on our full range cost volume, while other components remains unchanged.
The 3D-CNN is applied on the 1-channel correlation volume computed according to Eq.\eqref{eq: agg_correlation}.
The 3D-CNN then outputs the logit for each depth hypothesis, and the logits per pixel are further normalized by the soft-max function.
This provides more global matching information to CHOSEN.

Compared to the correlation volume in the baseline, the 3D-CNN and the normalization allow CHOSEN benefit from the higher quality matching signal.
Thus we can use just $\mathcal{L}_{\text{cl}}$ loss to supervise the whole pipeline, without the loss $\mathcal{L}_{\text{fm}}$ in our baseline MVS.
We found that this model produces performs much better on inputs with large texture-less regions, such as from the ETH3D datasets shown in Figure \ref{fig:with_3dcnn}.
We thus believe our formulation of CHOSEN will present mutual benefit with many of the current MVS frameworks that focus on cost volume and feature quality.



\section{Limitations and Conclusion}
We have demonstrated that CHOSEN is simple yet effective for multi-view depth refinement.
One limitation of CHOSEN is that it becomes more expensive to sample and select from a large number of hypotheses at high resolution.
This is part of the reason why we settled at $\nicefrac{1}{4}$ resolution to arrive at a sweet spot of both performance and run time.
Another limitation is that we did not focus on bundle adjustment and joint filtering of depths for point cloud and surface reconstruction, which we believe should be a subject of its own interests, especially considering the proliferation of volumetric based methods such as \cite{zhang2021ners}.

{
    \small
    \bibliographystyle{ieeenat_fullname}
    \bibliography{reference}
}

\clearpage
\setcounter{page}{1}
\maketitlesupplementary
\paragraph{Motivation for the baseline design}
Our baseline MVS design has at least the following disadvantages:
\begin{itemize}
    \item the matching features do not receive gradients from the hypothesis selection module.
    \item simple feature cross correlation is used, which is very noisy.
    \item there is no cost volume processing to denoise the matching cost volume
\end{itemize}
This matching component is thus isolated from the selection module and has sub-optimal performance, struggling in case of large texture-less areas, perspective distortions, etc. An example evaluation result of the matching component only is given in Tab.\ref{tab:depth comparison}, row titled “baseline w/o CHOSEN”.
Even with the noisy matching cost and its noisy initial depth, the selection module is still powerful enough to yield high quality depth estimation that surpasses other state-of-art models in terms of depth and normal accuracy on DTU. Therefore, one should expect better performance if we have a cost volume that behaves better than our noisy cross correlation. We did NOT aim to select a canonical architecture for matching cost, which is the focus of many other frameworks. We did demonstrate that CHOSEN can recover good estimation even if the matching cost is very noisy.

\paragraph{Architecture details}
We describe the details of the U-Nets used in the baseline MVS in Tab.\ref{tab:arch}.
A block refers to a sequence of convolution-ReLU-batch\_norm layers, with specified kernel size and dilation rate.
The context features $\mathbf{c}$ for CHOSEN contain 64 and 48 channels at $\nicefrac{1}{8}$ and $\nicefrac{1}{4}$ resolutions.
These features will be reinitialized at a new cost volume config level with two 3x3-convolution layers.
When updating them during the iteration, we use two ConvGRUs, one operate at resized input features to half of the size and one operate at full-size input features, and combine them through addition.
The ranking module has four 1x1-convolution layers, with intermediate channel size 64.
Our baseline MVS model is very small in terms of parameter count compared to other frameworks.

\begin{table*}[t!]
    \centering \footnotesize
    \begin{tabular}{cccccccc}
    \hline
    \multicolumn{8}{c}{\bf U-Net: matching and context feature extractors}\\
    \hline
    \textbf{Block name} & \textbf{Kernel sizes} & \textbf{Dilations} & \textbf{strides} & \textbf{Channels I/O} & \textbf{In res.} & \textbf{Out res.} & \textbf{Input name} \\
    \hline
    unet\_conv0  &  [4, 3]  &  [1, 1]  &  [1, 1]  &  3/16  &  1  &  1  &  $I_{\rm RGB}$ \\
    unet\_conv1  &  [4, 3]  &  [1, 1]  &  [2, 1]  &  16/32   & 1  &  1/2 & unet\_conv0 \\
    unet\_conv2  &  [4, 3]  &  [1, 1]  &  [2, 1]  &  32/48  &  1/2  &  1/4  &  unet\_conv1 \\
    unet\_conv3  &  [4, 3]  &  [1, 1]  &  [2, 1]  &  48/48  &  1/4  &  1/8 & unet\_conv2\\
    unet\_conv4* &  [4, 3]  &  [1, 1]  &  [2, 1]  &  48/48 &  1/8  &  1/16  & unet\_conv3 \\
    unet\_conv5* &  [3, 3]  &  [1, 1]  &  [2, 1]  &  96/48  &  1/16  &  1/8  & unet\_conv3, unet\_conv4 \\
    unet\_conv6* &  [3, 3]  &  [1, 1]  &  [2, 1]  &  96/48  &  1/8  &  1/4  & unet\_conv2, unet\_conv5 \\
    unet\_conv7* &  [3, 3]  &  [1, 1]  &  [2, 1]  &  80/32  &  1/4  &  1/2  & unet\_conv1, unet\_conv6 \\
    unet\_conv8* &  [3, 3]  &  [1, 1]  &  [2, 1]  &  48/16  &  1/2  &  1  & unet\_conv0, unet\_conv7 \\ 
    \hline
    \end{tabular} 
    \caption{Network Architecture details for the matching and context U-Nets in the baseline MVS.}
    \label{tab:arch}
\end{table*}

We explore a very simple extended setting to our baseline MVS, where we apply an MVSNet \cite{yao2018mvsnet} style 3D-CNN on our full range cost volume, while other components remains unchanged.
The 3D-CNN is applied on the 1-channel correlation volume computed according to Eq.\eqref{eq: agg_correlation}.
The 3D-CNN then outputs the logit for each depth hypothesis, and the logits per pixel are further normalized by the soft-max function.
This provides more global matching information to CHOSEN.

Compared to the correlation volume in the baseline, the 3D-CNN and the normalization allow CHOSEN benefit from the higher quality matching signal.
Thus we can use just $\mathcal{L}_{\text{cl}}$ loss to supervise the whole pipeline, without the loss $\mathcal{L}_{\text{fm}}$ in our baseline MVS.
We found that this model produces performs much better on inputs with large texture-less regions, such as from the ETH3D datasets shown in Figure \ref{fig:with_3dcnn}.
We thus believe our formulation of CHOSEN will present mutual benefit with many of the current MVS frameworks that focus on cost volume and feature quality.

\paragraph{Additional results on MultiFace}
We found MultiFace dataset a good testing ground (good calibration, high quality images) and close to the application scenarios our method is originally designed for.
In Fig. we provide more qualitative results. Note that the meshes provided in MultiFace dataset are too coarse to be ground truth for meaningful quantitative evaluation.

\begin{figure*}[t!]
\centering
\subfigure{\includegraphics[width=0.16\textwidth]{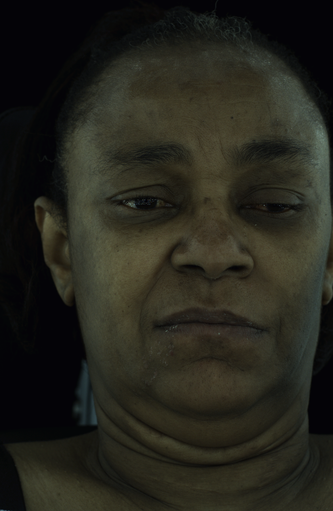}}
\subfigure{\includegraphics[width=0.16\textwidth]{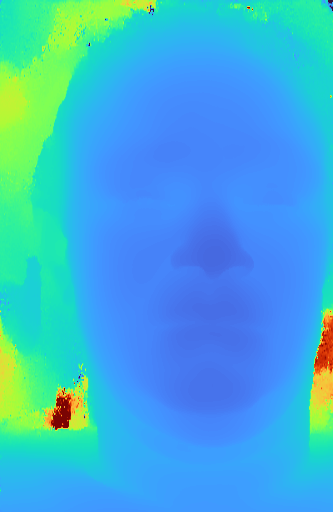}}
\subfigure{\includegraphics[width=0.16\textwidth]{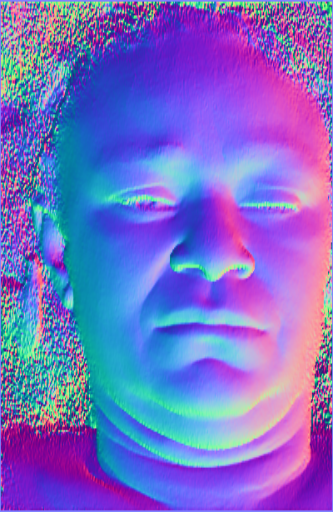}}
\subfigure{\includegraphics[width=0.16\textwidth]{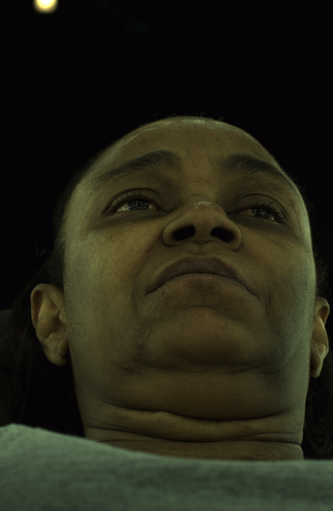}}
\subfigure{\includegraphics[width=0.16\textwidth]{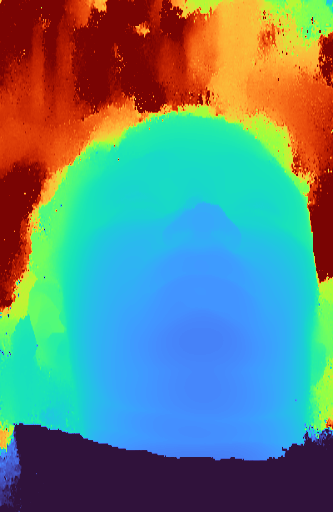}}
\subfigure{\includegraphics[width=0.16\textwidth]{./content/figures/wild_test/mf_12_color.png}}
\caption{Direct application of our DTU \cite{AanaesJVTD16_DTU} + BlendedMVS \cite{yao2020blendedmvs} trained baseline model on instances from MultiFace \cite{wuu2022multiface}. }
\label{fig:multiface test 1}
\end{figure*}

\begin{figure*}[t!]
\centering
\subfigure{\includegraphics[width=0.16\textwidth]{./content/figures/wild_test/mf_3_color.png}}
\subfigure{\includegraphics[width=0.16\textwidth]{./content/figures/wild_test/mf_3_depth.png}}
\subfigure{\includegraphics[width=0.16\textwidth]{./content/figures/wild_test/mf_3_normal.png}}
\subfigure{\includegraphics[width=0.16\textwidth]{./content/figures/wild_test/mf_4_color.png}}
\subfigure{\includegraphics[width=0.16\textwidth]{./content/figures/wild_test/mf_4_depth.png}}
\subfigure{\includegraphics[width=0.16\textwidth]{./content/figures/wild_test/mf_4_normal.png}} \\ \vspace{-10pt}
\subfigure{\includegraphics[width=0.16\textwidth]{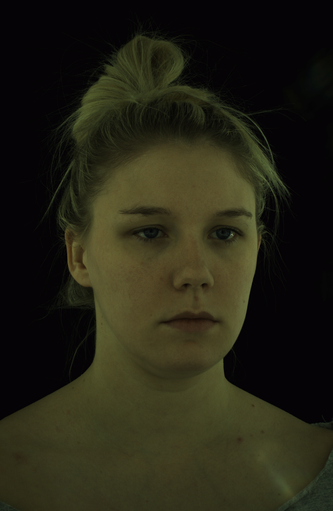}}
\subfigure{\includegraphics[width=0.16\textwidth]{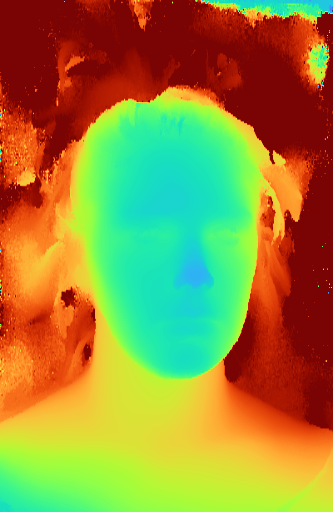}}
\subfigure{\includegraphics[width=0.16\textwidth]{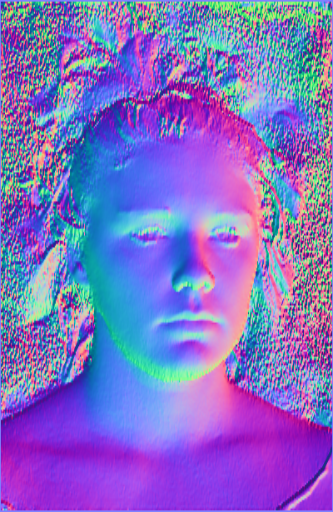}}
\subfigure{\includegraphics[width=0.16\textwidth]{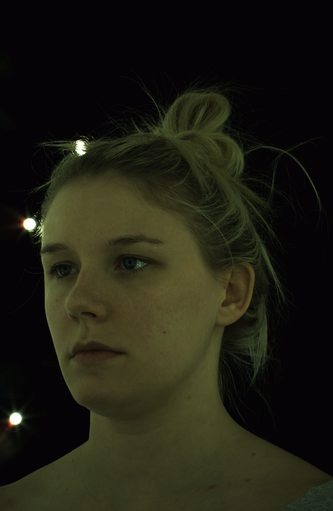}}
\subfigure{\includegraphics[width=0.16\textwidth]{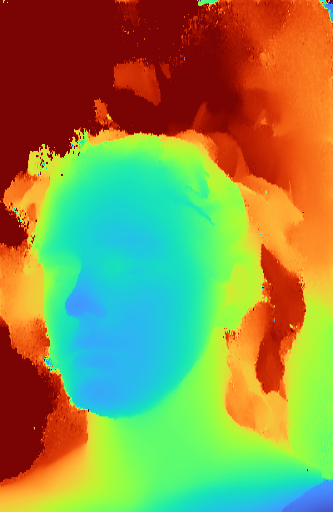}}
\subfigure{\includegraphics[width=0.16\textwidth]{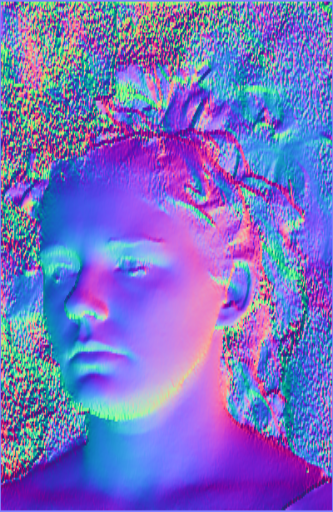}} \\ \vspace{-10pt}
\subfigure{\includegraphics[width=0.16\textwidth]{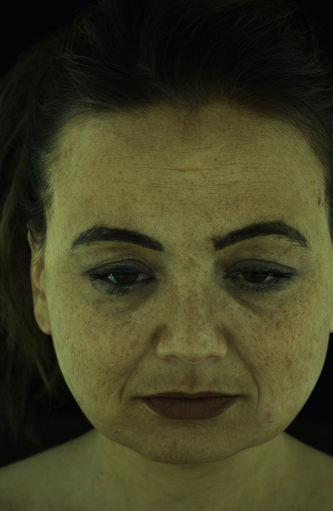}}
\subfigure{\includegraphics[width=0.16\textwidth]{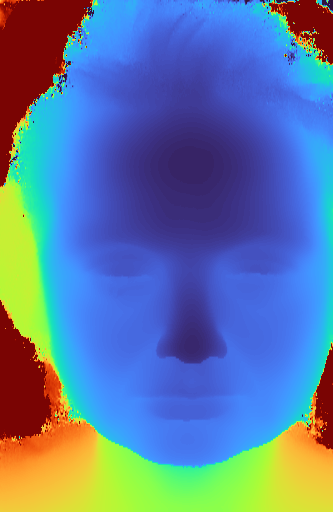}}
\subfigure{\includegraphics[width=0.16\textwidth]{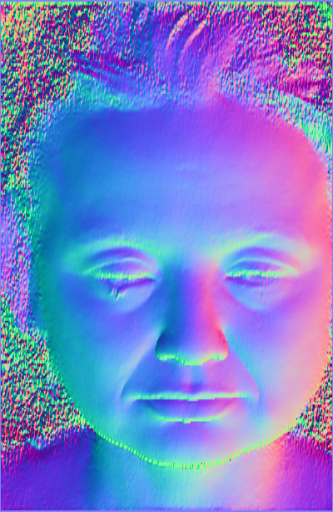}}
\subfigure{\includegraphics[width=0.16\textwidth]{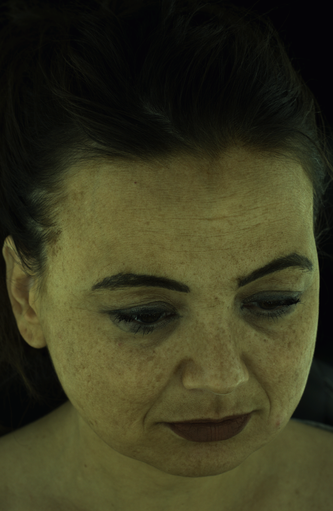}}
\subfigure{\includegraphics[width=0.16\textwidth]{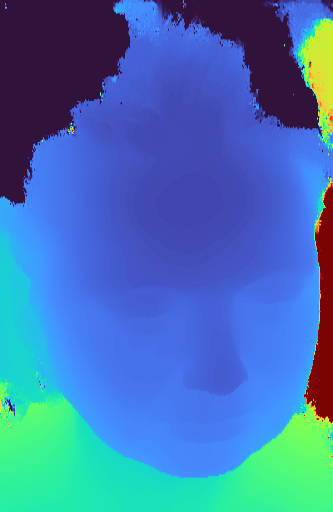}}
\subfigure{\includegraphics[width=0.16\textwidth]{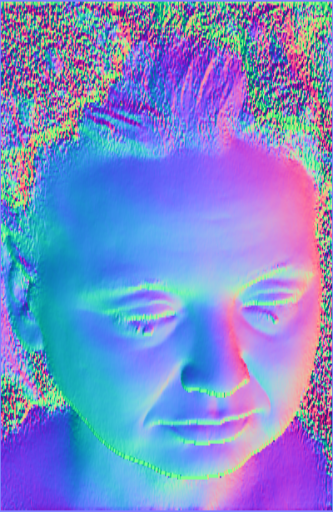}} \\ \vspace{-10pt}
\subfigure{\includegraphics[width=0.16\textwidth]{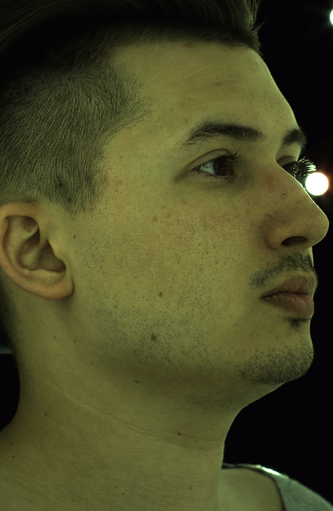}}
\subfigure{\includegraphics[width=0.16\textwidth]{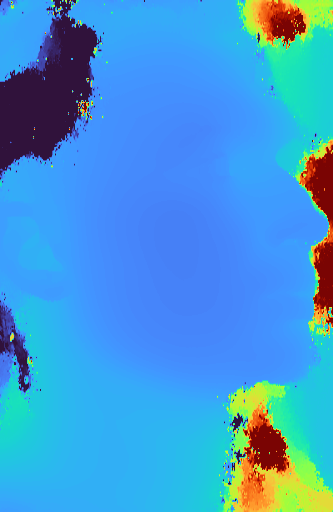}}
\subfigure{\includegraphics[width=0.16\textwidth]{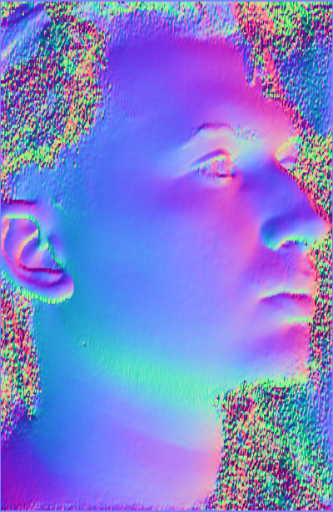}}
\subfigure{\includegraphics[width=0.16\textwidth]{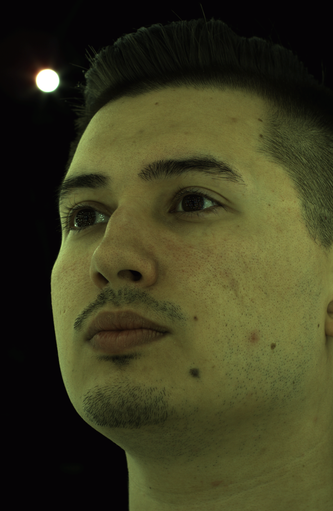}}
\subfigure{\includegraphics[width=0.16\textwidth]{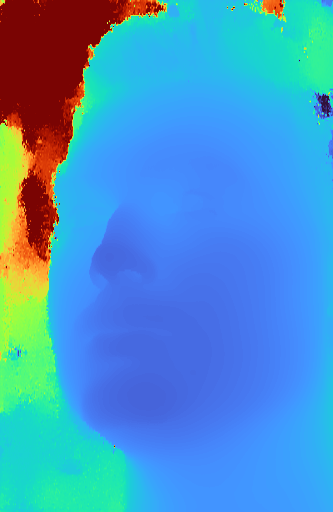}}
\subfigure{\includegraphics[width=0.16\textwidth]{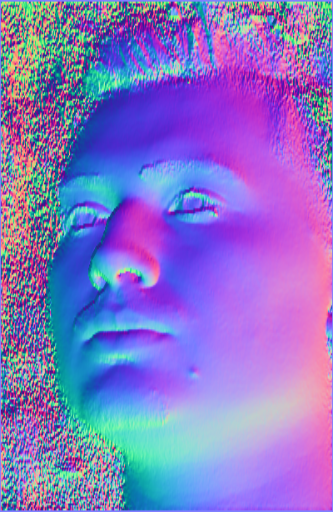}}
\caption{Direct application of our DTU \cite{AanaesJVTD16_DTU} + BlendedMVS \cite{yao2020blendedmvs} trained baseline model on instances from MultiFace \cite{wuu2022multiface}, continued from Fig.\ref{fig:multiface test 1}}
\label{fig:multiface test 2}
\end{figure*}

\end{document}